\documentclass{article}

     \usepackage[preprint, nonatbib]{neurips_2020}
   \usepackage[square,numbers]{natbib}
   \usepackage{soul} 
   \usepackage[pdftex]{graphicx}
   \usepackage{algorithm}
    \usepackage{algpseudocode}
    \usepackage{wrapfig}

\usepackage[utf8]{inputenc} 
\usepackage[T1]{fontenc}    
\usepackage{hyperref}       
\usepackage{url}            
\usepackage{booktabs}       
\usepackage{amsfonts}       
\usepackage{nicefrac}       
\usepackage{microtype}      

\title{Learning aligned embeddings for semi-supervised word translation using Maximum Mean Discrepancy}

\author{%
  Antonio H. O. Fonseca \\
  Interdepartmental Neuroscience Program\\
  Yale School of Medicine and Dept. of Computer Science \\
  New Haven - CT, United States of America\\
  \texttt{antonio.fonseca@yale.edu} \\
   \AND
   David van Dijk \\
   Yale School of Medicine and Dept. of Computer Science \\
   New Haven - CT, United States of America \\
   \texttt{david.vandijk@yale.edu} \\
}

\begin{document}

\maketitle

\begin{abstract}

Word translation is an integral part of language translation. In machine translation, each language is considered a domain with its own word embedding. The alignment between word embeddings allows linking semantically equivalent words in multilingual contexts. Moreover, it offers a way to infer cross-lingual meaning for words without a direct translation. Current methods for word embedding alignment are either supervised, i.e. they require known word pairs, or learn a cross-domain transformation on fixed embeddings in an unsupervised way. Here we propose an end-to-end approach for word embedding alignment that does not require known word pairs. Our method, termed Word Alignment through MMD (WAM), learns embeddings that are aligned during sentence translation training using a localized Maximum Mean Discrepancy (MMD) constraint between the embeddings. We show that our method not only out-performs unsupervised methods, but also supervised methods that train on known word translations. 

\end{abstract}

\section{Introduction}

Word translation can be challenging because often there exists no single word translation capable of encapsulating the meaning of the original word \cite{ahluwalia2019pristine}. For example, the Brazilian Portuguese word \textit{cafuné} means "to run your fingers through the hair of someone you love". Such a rich physical expression of human affection and love cannot be translated by a single English word, but can be approximated by the combined use of words that represent the act. In machine translation, the vocabulary of a language is often represented by a word embedding which is generally specific to the language \cite{ruder2019survey}. A word embedding provides a metric space that represents semantics where words with similar meanings are closer to each other. Consequently, word translation can be achieved by finding a mapping function between the word embedding of two languages \cite{jansen2017word}. This can be done by either learning a transformation \cite{mikolov2013exploiting, alvarez2018gromov, gaddy2016ten} or learning two embeddings that are aligned \cite{garg2019jointly}, i.e. are part of the same space.
This has been done in supervised \cite{mikolov2013exploiting}, semi-supervised \cite{garg2019jointly, gaddy2016ten}, and unsupervised \cite{alvarez2018gromov} manners. Although several methods for word translation have been proposed, most require large numbers of provided word pairs (supervised) to achieve a good performance \cite{mikolov2013exploiting, gaddy2016ten}, or (for unsupervised methods) may lack sufficient references to achieve a good quality alignment \cite{alvarez2018gromov}. In this work, we present a new method that learns two aligned embeddings, thus providing accurate and semantically meaningful word translation, without the need for known (supervised) word pairs. Our method, termed WAM, is based on a localized Maximum Mean Discrepancy (MMD) loss that is added to a sentence translation task (e.g. Transformer \cite{vaswani2017attention}) and aims to align the embeddings by minimizing their distributional distance. We show that our method outperforms other methods that are both supervised (with provided word translation pairs) and unsupervised.



\section{Related work}
The problem of word alignment has been tackled with a variety of methods. Approaches for word alignment can generally be classified into methods based on statistical machine translation or neural machine translation. Here we briefly discuss these approaches.

\subsection{Statistical machine translation}
The geometry of the word embedding for a language is highly dependent on the distributed representation of words. Therefore, the geometries of word embeddings for two distinct languages should have a relative degree of similarity. Methods such as the Gromov-Wasserstein Alignment \cite{alvarez2018gromov} have exploited this assumption to achieve unsupervised word embedding alignment. \citet{alvarez2018gromov} proposes to align two embeddings via optimal transport based on the Gromov-Wasserstein distances between pairs of words within each embedding, i.e. distances across embeddings rather than the distance between words. The solution to the optimal transport of Gromov-Wasserstein distances is an optimal coupling matrix between the embeddings which yields minimal discrepancy cost. Despite the potential benefits of this approach, it has several major drawbacks, such as being an expensive computation for large vocabulary sizes and having general poor performance. The poor performance (as we show in section \ref{Analysis}) may stem from the fact that the optimal coupling, which aligns the distributions as a whole, may not correspond to the correct alignment.

\subsection{Neural Machine Translation}
In machine translation, neural approaches have shown superior accuracy over purely statistical methods. In particular, architectures such as the Transformer \cite{vaswani2017attention}, which makes use of an attention mechanism, have shown far better translation performance due to their capability of attending to the context of the words rather than literal translation \cite{vaswani2017attention}. However, the probability distribution given by the attention mechanism may not necessarily allow for inference on word alignment between vocabularies \cite{koehn2017six}, thus requiring the use of statistical approaches for word alignment \cite{alvarez2018gromov}. To address this problem, \citet{garg2019jointly} proposed to train a Transformer in a multi-task framework. Their multi-task loss function combines a negative log-likelihood loss of the Transformer (relative to the token prediction based on past tokens) with a conditional cross-entropy loss defined by the Kullback-Leibler (KL) divergence. In their approach, they used the attention probabilities from the penultimate attention head layer as labels for their supervised algorithm, thus dispensing the need for an annotated word alignment. The divergence between the attention probabilities of one arbitrarily chosen alignment head and the labeled alignment distribution is minimized as a KL divergence optimization problem \cite{garg2019jointly}.



\section{Proposed method}
Our method, termed Word Alignment through Maximum Mean Discrepancy (WAM) achieves word translation by using the Transformer model \cite{vaswani2017attention} in combination with a localized Maximum Mean Discrepancy (MMD) loss. Briefly, while the embedding for each language is learned during the Transformer training for a sentence translation task, the MMD loss applies local constraints on paired sentences in order to learn a pair of embeddings that are locally, and as a result globally, aligned. The end result is an accurate word alignment between languages while maintaining sentence translation performance. Here we describe the details of our method.

\subsection{Word and sentence alignment}
\label{word_align_back}
The alignment between the words of two languages can be thought of as a general alignment problem. In this formulation, the assumption is that the same sentence expressed in two different languages is still linked by semantics and therefore contains a significant shared underlying structure \cite{wang2009general}. Thus, using the notation from \citet{ruder2019survey}, the word embedding matrix learned for a language $l$ with a vocabulary $V^l$ and dimension $d_{model}$ can be described as $ X^{l} \in \mathbb{R}^{V^l \times d_{model}}$. Consequently, the $j$-th word out of $n$ words in the $i$-th sentence out of $m$ sentences is defined as $\{x^{l}_{i,j} |  i \in \mathbb{Z}: i \in [1..m], j \in \mathbb{Z}: j \in [1..n] \}$. The alignment task consists of minimizing the distance between the words from a source sentence $x^{S}_{i,j}$ and a target sentence $x^{T}_{i,j}$ based on their meaning.

\subsection{Transformer}
\label{Transformer}
The Transformer is a model architecture based purely on an attention mechanism to compute the contextual relationship between source and target languages \cite{vaswani2017attention}. This model has gained significant popularity due to its performance on neural machine translation. The architecture of the Transformer is composed of a stack of encoders and decoders. Each encoder is composed of a self-attention layer and a feed-forward neural network. The self-attention computes a weighted sum over all the embedded input words in the source sentence, thus providing information about the context. The output of the self-attention layer is fed to the feed-forward network. Inputs from the encoder are sent to the decoder, which contains both the self-attention and feed-forward network, but with an attention layer in between them. This attention layer receives the inputs from the encoder and it is responsible for learning how to link word representations from the source language to a representation of the target language.

Each word is represented in an embedded space of size $d_{model}=512$. This embedding consists on learning a weight matrix $ W_{embed} \in \mathbb{R}^{V^l \times d_{model}}$, thus returning a $ x^{l}_{i,j} \in \mathbb{R}^{1 \times d_{model}}$. In the self-attention layer, the input  $x^{l}_{i,j}$ is used to compute the importance of other words relative to its own predicted output (query $Q_{i,j}=W_{Q}x_{i,j}$), its importance on the output of other words (key $K_{j}=W_{K}x_{i,j}$) and the output as a weighted sum of all words (value,  $V_{j}=W_{V}x_{i,j}$). These are learned linear transformations of the input $x_{i,j}$, wherein $W_{Q}\in \mathbb{R}^{d_{model} \times d_{K}}$, $W_{K} \in \mathbb{R}^{d_{model} \times d_{K}}$, $W_{V} \in \mathbb{R}^{d_{model} \times d_{V}}$. The attention to the $j$-th token relative to other tokens is given by

\begin{equation}
    Attention(Q_{j},K_{j},V_{j}) = softmax(\frac{Q_{j}K_{j}}{\sqrt{d_{K}}})V_{j}
\end{equation}

This operation is computed in parallel by 8 attention heads. Their individual outputs are linearly combined to generate a single attention value relative to the $j$-th token.

The loss function used for the Transformer is based on a KL-divergence loss of the smoothed labels with $\epsilon=0.1$, such as done by \citet{vaswani2017attention}. This replaces the conventional one-hot distribution by a distribution with relative confidence on the correct label and smoothed mass over the vocabulary \cite{szegedy2015rethinking}. 




\subsection{Global Word Alignment via Local MMD}
\label{MMD}
Previously proposed methods aim to perform embedding alignment by assuming that the underlying structure of the embedding is conserved across languages \cite{alvarez2018gromov,mikolov2013exploiting}. Although it may be locally true, there is no guarantee that such an assumption holds to be true for the entire embedding. On the other hand, a sentence represented in two different languages should convey the same meaning and, consequently, similar word usage within each language. Based on this assumption, we propose to achieve word alignment by computing semantic similarity between pairs of sentences via a sentence based distributional distance. During the training process, the Transformer learns to embed the words of a language based on their context, thus generating an embedding $ X^{l} \in \mathbb{R}^{V^l \times d_{model}}$. Therefore, the semantic similarity between the words can be computed in terms of their distances in the embedding. Since we cannot make inferences directly on word distances (as we do not make use of known word pairs), we compute the distance between distributions of words in the sentences by using Maximum Mean Discrepancy (MMD) as a distance measure \cite{gretton2012kernel}. We thus compute a localized MMD, as it involves the distribution of words within a sentence (as opposed to the distribution of words within the whole language vocabulary). This results in the alignment of distributions of a pair of sentences between languages. We do not compute MMD globally, i.e. on the whole vocabulary, as this may give a distributional alignment that lacks the correct word alignment. However, since we minimize MMD on many sentence pairs, the result is an embedding that is also globally (correctly) aligned.

The MMD is defined in terms of particular function spaces that present a difference in distributions. Unlike a distributional distance such as KL-divergence, the MMD can be formulated directly on the empirical expectation of the samples and dispenses density estimation. This grants a dimension-independent property and preserves information about the original distributions. The empirical expectation of the MMD can be written as follows:


\begin{equation}
    MMD(X^{S},X^{T}) = \frac{1}{m}\sum^{m}_{i=1}\left\Vert\frac{1}{n}\sum^{n}_{j=1}x_{i,j}^{S} - \frac{1}{n}\sum^{n}_{j=1}x_{i,j}^{T} \right\Vert^{2}_{\mathcal{H}}
\end{equation}

where $x_{i,j}^{S}$ and $x_{i,j}^{T}$ denote $i$-th sentence and $j$-th word for source ($S$) and target ($T$) languages, respectively. The operation is done in a Hilbert space ${\mathcal{H}}$. 

We rewrite the MMD in terms of a kernel $k$ as follows:

\begin{equation}
    MMD(X^{S},X^{T}) = \frac{1}{m}\sum^{m}_{i=1}\left(\frac{1}{n^2}\sum^{n}_{a,b=1}k(x_a^{S},x_b^{S}) + \\ \frac{1}{n^2}\sum^{n}_{a,b=1}k(x_a^{T},x_b^{T}) - \\
    \frac{2}{n^2}\sum^{n}_{a,b=1}k(x_a^{S},x_b^{T}) \right)
\end{equation}


For our application, we choose a Gaussian RBF kernel. To overcome the challenge of finding a $\sigma$ that is appropriate for the entirety of the dataset, we use a multi-scale kernel approach, which is defined as follows:

\begin{equation}
    k(x,x') = \sum^{\gamma_{n}}_{\sigma=\gamma_{i}}\exp\left( \frac{-1}{2\sigma^2} \Vert{x-x'}\Vert^2 \right)
\end{equation}

where $\{\gamma_{i}=10^{i} | i \in \mathbb{Z}: i \in [-3,2]\}$.

\begin{figure}
  \centering
    \includegraphics[width=0.9\linewidth]{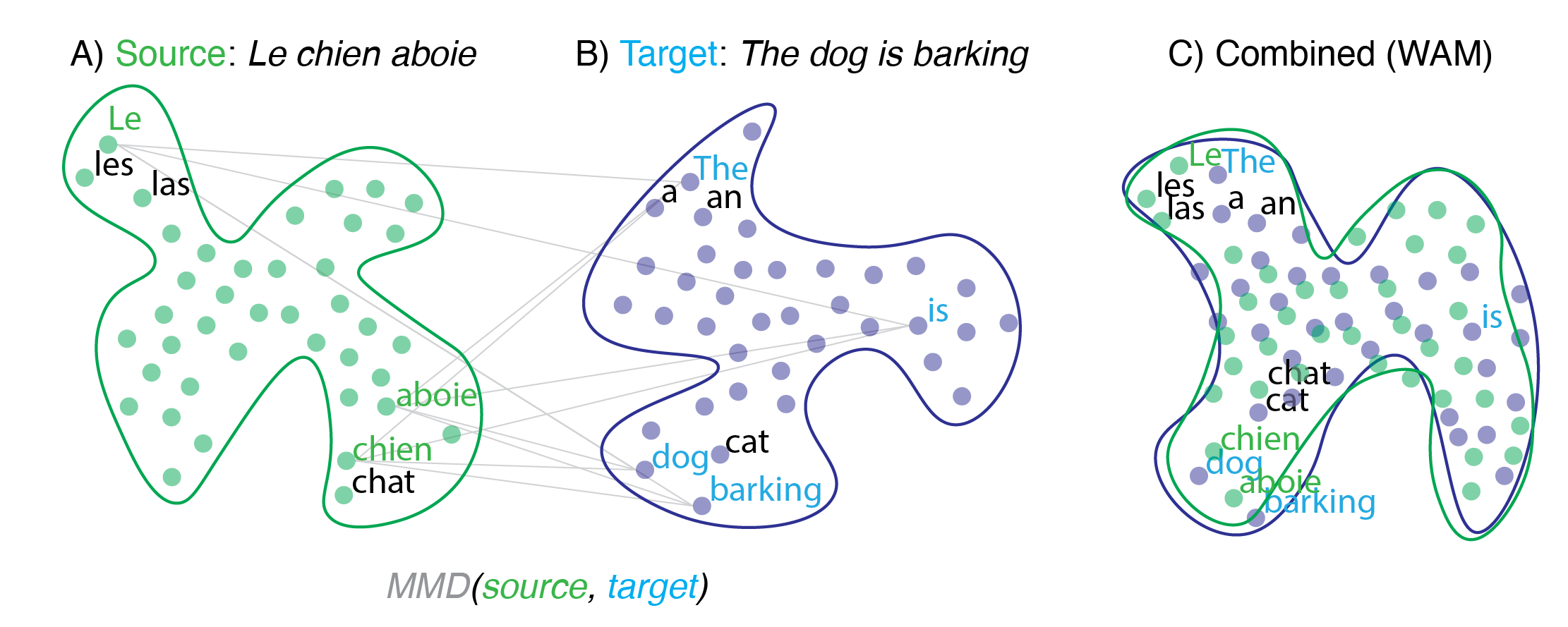}
  \caption{\textbf{Representation of the alignment process with WAM}. The computation performed by WAM consists on minimizing the weighted sum of the Transformer and the MMD losses. At each training step, the Transformer minimizes the loss regarding token prediction based on context (sec. \ref{Transformer}) while the MMD minimizes the distance between sets of tokens of paired sentences (sec. \ref{MMD}). For illustration of the process, the diagram shows the tokens from the source sentence (\textit{'Le'},\textit{'chien'} and \textit{'aboie'}) in the French embedding (A) and the tokens from the target sentence (\textit{'The'},\textit{ 'dog'}, \textit{'is'} and \textit{'barking'}) in the English embedding (B) becoming aligned without losing the structural semantic embedding giving by the Transformer (C). } 
  \label{Figure0_Method}
\end{figure}

Therefore, we expect the MMD to have small values when the distribution of the word sets are similar (i.e., sentences with similar semantic content) and large for disparate distributions. The global loss of one optimization step is computed as a weighted sum of the Transformer loss ($L_T$) and the MMD loss ($L_M$) (Figure \ref{Figure0_Method} and Algorithm \ref{WAM_algo}). For our tests, we multiply the MMD loss by 10 in order to have a similar magnitude to the Transformer loss. The global loss ($L$) is minimized with respect to the Transformer parameter $\theta$, where $\theta$ is defined as all the learnable parameters of the Transformer  (see sec. \ref{Transformer}).

In the following sections, we show how our method contributes to word embedding alignment when compared to other approaches.

\begin{algorithm}
  \caption{WAM computation for Word Embedding Alignment}\label{WAM_method}
  \begin{algorithmic}[1]
    \Require{Initial source and target embeddings $X^S$, $X^T$. Weights $\theta$}
    \Procedure{WAM}{$X^S,X^T$}\Comment{alignment of $X^S$, $X^T$}
      \While{not converged}\Comment{We have the answer if r is 0}
        \State{$a \gets x^{S}_{i,j} | x^{S}_{i,j} \subseteq X^S$  \Comment{Select subset sentences}}
        \State $b \gets x^{T}_{i,j} | x^{T}_{i,j} \subseteq X^T$
        \State $L_{T} \gets Transformer_{\theta} (a,b)$ \Comment{Compute the Transformer loss (sec. \ref{Transformer})}
        \State $L_M \gets 0$
        \For{$i \gets 1$ to $N$} \Comment{Iterate through $N$ sentences}
            \State $X_a,X_b \gets Embed_{\theta}(a_i,b_i)$ \Comment{Word embeddings from Transformer}
            \State $L_M \gets L_M + MMD(X_{a},X_{b})$ \Comment{MMD loss (sec. \ref{MMD})}
        \EndFor{}
        \State $L \gets L_T + L_M$
        \State $backpropagation(L_{\theta})$ \Comment{Minimize loss wrt $\theta$ (sec. \ref{MMD})}
      \EndWhile\label{euclidendwhile}
    \EndProcedure
  \end{algorithmic}
\label{WAM_algo}
\end{algorithm}

\section{Experiments}
\subsection{Data and model setup}
\label{Dataset and setup}
In our experiments, we used the French-English parallel corpus of the Europarl-v7 \cite{koehn2005europarl} dataset for training. From the corpus, we selected the first 100k sentences to be used for development. We used spaCy library for tokenization. We filtered out sentences that were longer than 80 tokens in both languages and sentence pairs with a source/target length ratio greater than 1.5. This dataset was split in 90/10 for training and validation, respectively. For the test, we evaluated the performance on two French-English dictionaries: our manually built dictionary (1026 pairs of words) and the MUSE \cite{conneau2017word} dictionary (10872 pairs of words). Our dictionary contained commonly used words by both languages, such as pronouns, nouns, prepositions and verbs. The MUSE dictionary offers similar pairs in addition to alternative translations for the same words. For example, the word \textit{love} (English) has four entries in the MUSE dictionary as possible French translations: \textit{aime}, \textit{amour}, \textit{aimer} and \textit{love}.

In our experiments, we used the base Transformer setting with embedding size 512, 6 encoder and decoder layers, 8 attention heads and sinusoidal positional embedding. We used Adam optimizer with $\beta_1=0.9$, $\beta_2=0.98$ and $\epsilon=10^{-9}$. We varied the learning rate over the course of training, such that the learning rate increased linearly for the first 2000 training steps (warmup), then decreasing it proportionally to the inverse square root of the step number. The training was done in batches of 2500 tokens. We trained our models on a machine with a P100 GPU.

\subsection{Word alignment evaluation}
\label{Alignment evaluation}
To measure the quality of the alignments, we calculated the coefficient of determination ($R^2$) between the embeddings for a set of known word translation pairs. In specific, we computed this quantity per word pair and reported the average over all pairs. We perform this evaluation for word pairs from both our and the MUSE \cite{conneau2017word} dictionaries. A higher $R^2$ implies a better alignment and $R^2=1$ is a perfect alignment. In addition, we quantify model performance using a metric presented in \citet{alvarez2018gromov}, which is the accuracy for the $n$-nearest target neighbors to a target word. We compute this accuracy for 1, 5 and 10 nearest neighbors. These four metrics ($R^2$ and accuracies) are shown in Table \ref{Table performance}. For visualization of the embeddings, we reduced the dimensionality of the original embedding to a 2D space using UMAP \cite{mcinnes2018umap}, such as illustrated in Figure \ref{Figure1_Alignment}.

\subsection{Comparing methods}
\label{Comparing methods}
We compare our method to three state-of-the-art word alignment approaches: unsupervised Gromov-Wasserstein alignment \cite{alvarez2018gromov}, the neural-based Joint Learning alignment \cite{garg2019jointly}, and a supervised embedding alignment. We used a supervised alignment method that is based on a Transformer model, such as described in section \ref{Dataset and setup}, with a supervised word pair loss that is calculated between the embeddings of a set of known pair of "landmarks". As landmarks, we used the first half of the dictionary while keeping the second half for evaluation. In each epoch, we sampled 50\% of the given landmark pairs, computed their embedding and their distances as L2 norm. In this case, the total loss of the step was given as the sum of the Transformer loss and the landmark L2 norm loss.

\section{Analysis}
\label{Analysis}

\begin{figure}
  \centering
    \includegraphics[width=0.9\linewidth]{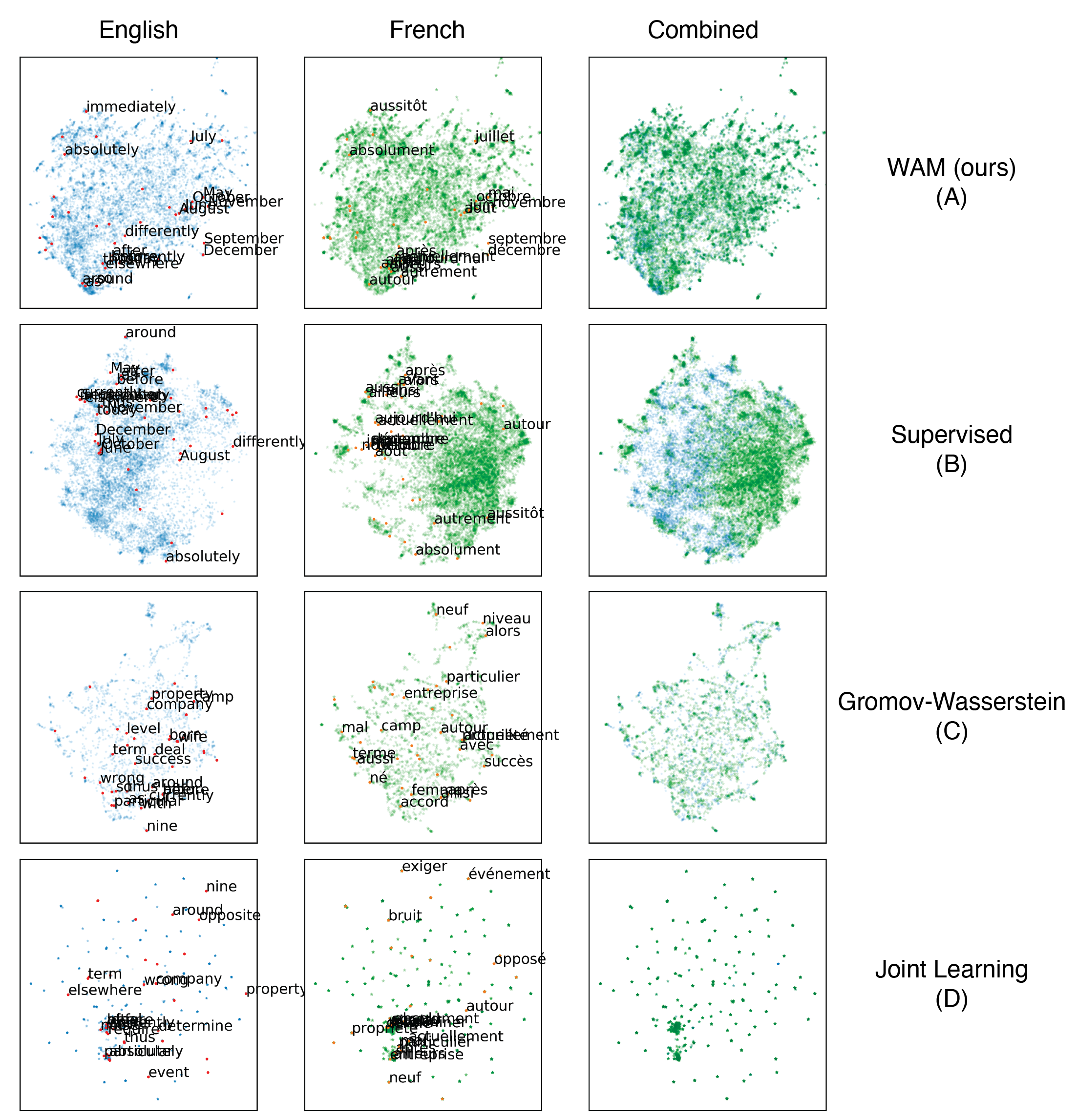}
  \caption{\textbf{2D visualization of the embeddings achieved by the different methods}. We reduced the dimensionality of the embeddings to 2D via UMAP \cite{mcinnes2018umap}. The resulting embedding for each method is displayed per row. In the English embedding (first column), words from the vocabulary list and our dictionary are shown in blue and red respectively. In the French embedding (second column), words from the vocabulary and our dictionary are represented in green and orange, respectively. The third column shows the overlay of the English and French embeddings.} 
  \label{Figure1_Alignment}
\end{figure}

Semi-supervised methods, such as the Joint Learning alignment \cite{garg2019jointly} and ours, were only trained on sentence pairs and not on word pairs. The supervised method was trained on both sentence pairs and word pairs. All methods were trained on sentence pairs from Europarl-v7. In addition, the Gromov-Wasserstein alignment method \cite{alvarez2018gromov} itself does not compute embeddings, thus requiring a pre-computed embedding as input. We trained the Gromov-Wasserstein alignment method on the French and English embeddings generated by our WAM method for the Europarl-v7 dataset (illustrated in Figure \ref{Figure1_Alignment}A) and on their own reported training set, which consisted of word embeddings for French and English trained with FastText on the Wikipedia dataset \cite{bojanowski2017enriching}.

The performances of the different methods were evaluated such as described in section \ref{Alignment evaluation}. The quantitative results are shown in Table \ref{Table performance}. Figure \ref{Figure1_Alignment} illustrates the alignment achieved by each method, wherein blue and green dots identify English and French words from the vocabulary (i.e, unique words among the 100k sentences from the Europarl-v7 dataset), respectively. The red and orange dots indicate French and English words, respectively, from our dictionary. 
 
\begin{table}
  \caption{Performance for various methods in a French to English translation task. The methods were tested on the MUSE \cite{conneau2017word} and our dictionary (sec. \ref{Dataset and setup}). The training on the Europarl-v7 dataset \cite{koehn2005europarl} was performed with 100k sentences and the training on Wikipedia was performed one the embedding provided by FastText \cite{bojanowski2017enriching}}
  \label{Table performance}
  \centering
  \begin{tabular}{lllllll}
  \toprule
  & & & \multicolumn{3}{c}{Nearest Neighbors} \\
  \cmidrule(r){4-6}
    Method  & Train & Test  &  1 NN &  5 NN  &  10 NN & $R^2$  \\
    \midrule
Gromov-Wasserstein \cite{alvarez2018gromov} & Wikipedia \cite{bojanowski2017enriching} &  ours   &  0.25 &  0.38 &  0.41 & 0.31 \\
Gromov-Wasserstein \cite{alvarez2018gromov} & Europarl-v7 \cite{koehn2005europarl}&  ours   &  0.00 & 0.00 &  0.00 & 0.00 \\
Joint Learning \cite{garg2019jointly}  & Europarl-v7 \cite{koehn2005europarl} & ours      &  0.02 &  0.03 &  0.04 & 0.14 \\
Supervised & Europarl-v7 \cite{koehn2005europarl} & ours & 0.01 &	0.02 & 	0.03 & 	0.03 \\
WAM (ours) & Europarl-v7 \cite{koehn2005europarl} & ours    &  \textbf{0.55} &  \textbf{0.64} &  \textbf{0.66} & \textbf{0.65} \\

Gromov-Wasserstein \cite{alvarez2018gromov} & Wikipedia \cite{bojanowski2017enriching} & MUSE \cite{conneau2017word} &  0.08 &  0.12 &  0.13 & 0.33 \\
Gromov-Wasserstein \cite{alvarez2018gromov} & Europarl-v7 \cite{koehn2005europarl}
&  MUSE \cite{conneau2017word}   &  0.00 & 0.00 &  0.00 & 0.00 \\
Joint Learning \cite{garg2019jointly}  & Europarl-v7 \cite{koehn2005europarl} & MUSE \cite{conneau2017word}      & 0.01 &	0.03&	0.03&	0.20  \\
Supervised & Europarl-v7 \cite{koehn2005europarl} & MUSE \cite{conneau2017word}  & 0.05& 0.06 & 0.06	& 0.16 \\
WAM (ours) & Europarl-v7 \cite{koehn2005europarl} & MUSE \cite{conneau2017word}  &  \textbf{0.36} &  \textbf{0.45} &  \textbf{0.46} & \textbf{0.37} \\
    \bottomrule
  \end{tabular}
\end{table}

First, we evaluated the performance of the supervised method. This model was trained as described in section \ref{Comparing methods}. The supervised method achieved low performance for all the four metrics in both dictionaries (Table \ref{Table performance}). This result may be expected as correctly aligning words pairs from a training set does not guarantee generalization to the unseen words. In this setup, the distance between the landmark word pairs appears to be quickly minimized (to zero distance) while the Transformer is still shaping the embeddings around the landmarks, which at this point act as anchors. The result of this is low performance for the unseen part of the dictionary (details about the split in section \ref{Comparing methods}) while keeping a constantly high coefficient of determination ($R^2=1$) for the seen part of the dictionary (data not shown here). Figure \ref{Figure1_Alignment}B shows the resulting alignment between the embeddings with the supervised method for our dictionary (both seen and unseen words). The embeddings present a substantial amount of misalignment, as the quantification also indicates (Table \ref{Table performance}). These results show that supervised word embedding alignment gives poor global alignment. Better performance may be achieved by using a bigger set of landmarks, but this also makes the approach impractical as the goal is to learn a generalizable alignment and not to memorize all known word translation pairs.

Next, we evaluated the performance of the Gromov-Wasserstein method \cite{alvarez2018gromov}. This method presented poor overall performance when trained on our embedding, which motivated us to evaluate their performance using their own reported training set. We observed satisfactory performance for the latter setup in both dictionaries (Table \ref{Table performance}). Their achieved alignment for our dictionary is illustrated in Figure \ref{Figure1_Alignment}C. Based on these results, we concluded that the Gromov-Wasserstein method may achieve distributional alignment when trained on very large datasets such as the FastText on the Wikipedia dataset \cite{bojanowski2017enriching}, but their word-to-word mapping is suboptimal as indicated by labeled landmarks in Figure \ref{Figure1_Alignment}C and Table \ref{Table performance}. In other words, their method (visually) appears to align the distributions however may not actually give the correct word alignment.

\begin{figure}
  \centering
    \includegraphics[width=1\linewidth]{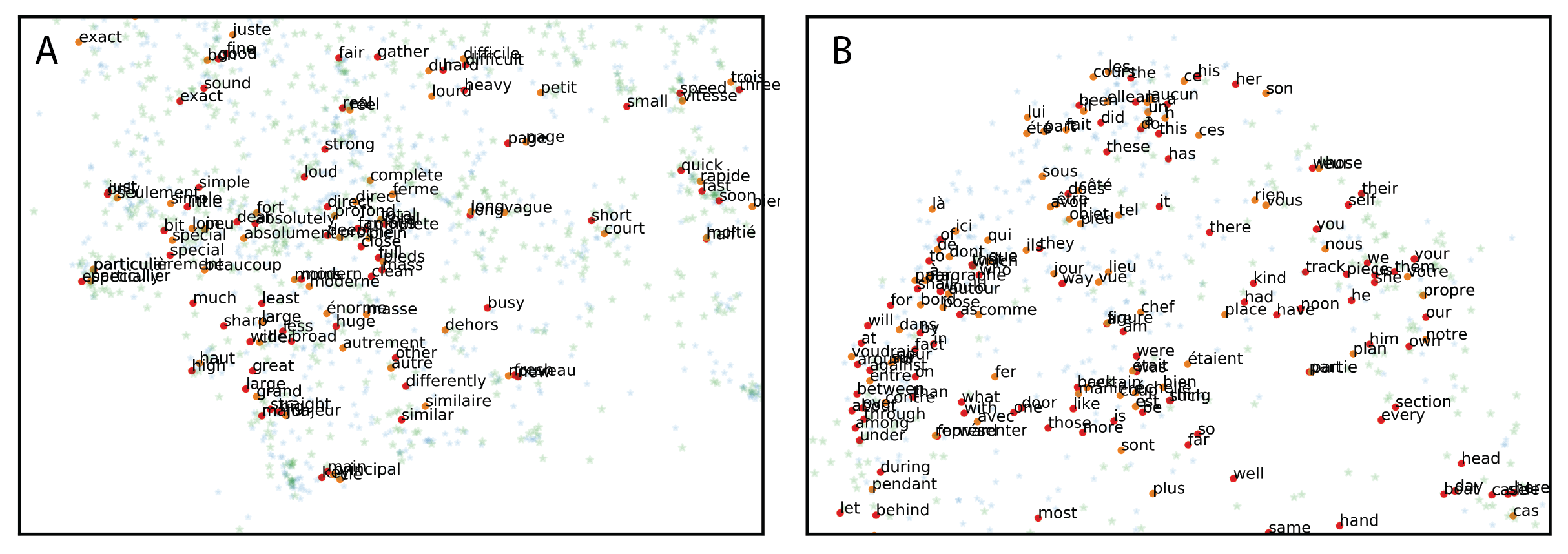}
  \caption{\textbf{Zoomed in details of the combined embedding obtained with our method}. The blue and green stars represent words from English and French vocabularies, respectively. The red and orange circles indicate known word pairs in English and French, respectively, and are shown with their labels. Panel A and B show two different regions of the same embedding.}
  \label{Figure2_ZoomIn}
\end{figure}

We then evaluated the performance of the Joint Learning method \cite{garg2019jointly}. For a better comparison with our method, we trained their model with a vanilla Transformer, which is essentially the same architecture as the Transformer we use (see section \ref{Dataset and setup}). As shown in Table \ref{Table performance}, this method showed a performance comparable to the supervised learning approach based on the four reported metrics, suggesting their low fit for the word-to-word translation tasks. Their resulting embedding is shown in Figure \ref{Figure1_Alignment}D, where a large number of clusters can be appreciated. As described in their work, the Joint Learning Alignment focuses on linking the words from a source sentence to a target sentence in order to provide the best word-to-word translation \cite{garg2019jointly}. This implies that their method does not aim for vocabulary alignment like we do, although it can still be achieved by giving single words as inputs to their trained model.

Finally, we evaluated the performance of WAM, our proposed semi-supervised approach that does not require known word pairs. WAM showed superior performance in all the four metrics used for both dictionaries (Table \ref{Table performance}), thus indicating good generalization capacity. The resulting embedding presented a (visually) nearly perfect alignment (Figure \ref{Figure1_Alignment}A) while preserving the semantic arrangement given by the Transformer, as illustrated by the combined embedding (zoomed in) detail (Figure \ref{Figure2_ZoomIn}).

\section{Conclusion}
In this work, we presented WAM (Word Alignment through Maximum Mean Discrepancy), an end-to-end approach for word embedding alignment. We have shown that our method outperforms other supervised, unsupervised, and semi-supervised methods in several metrics that measure word alignment quality between languages. Our method is based on the Transformer model for sentence translation and uses a novel localized Maximum Mean Discrepancy (MMD) loss, which allows for a semi-supervised matching between word distributions while still learning the semantic embedding. Our method provides an accurate mapping between the embedding of two languages that is not uniquely dependent on the existence of a target word. This feature allows the semantic inference of a source word by inspection of the immediate neighbors of the estimated target location. This is an important step towards bridging a semantic gap between languages with great potential for cultural enrichment and a better understanding between different cultures. 

\section{Broader Impact}
Machine translation has deep societal impact as it allows different cultures to communicate. Word translation is an important part of language translation and presents challenges when there exists no single perfect translation for a word. As such, the inaccuracy of a translation model can cause problems in communication and result in misinterpretation. Many models fail to perform correct word translation due to the difficulty of finding the proper mapping between the domains of two languages. Our method achieves high accuracy word translation via learning of an aligned metric space for two languages. This allows the estimation of a target embedding coordinate even when no single word is available by providing a semantic description using several of the nearest words. This, in general, can potentially give a richer and more meaningful translation when neighboring words are taken into account. One caveat with our method, as is the case for all machine translation models, is that its translations may be biased when it is trained on text from a specific domain and thus machine translations (including word translations) should always be looked at critically.

\bibliography{Fonseca_VanDijk.bib}

\begin{thebibliography}{16}
\providecommand{\natexlab}[1]{#1}
\providecommand{\url}[1]{\texttt{#1}}
\expandafter\ifx\csname urlstyle\endcsname\relax
  \providecommand{\doi}[1]{doi: #1}\else
  \providecommand{\doi}{doi: \begingroup \urlstyle{rm}\Url}\fi

\bibitem[Ahluwalia et~al.(2019)Ahluwalia, Coari, and
  Brock]{ahluwalia2019pristine}
M.~Ahluwalia, B.~Coari, and B.~Brock.
\newblock Pristine sentence translation: A new approach to a timeless problem.
\newblock \emph{SMU Data Science Review}, 2\penalty0 (2):\penalty0 4, 2019.

\bibitem[Alvarez-Melis and Jaakkola(2018)]{alvarez2018gromov}
D.~Alvarez-Melis and T.~S. Jaakkola.
\newblock Gromov-wasserstein alignment of word embedding spaces.
\newblock \emph{arXiv preprint arXiv:1809.00013}, 2018.

\bibitem[Bojanowski et~al.(2017)Bojanowski, Grave, Joulin, and
  Mikolov]{bojanowski2017enriching}
P.~Bojanowski, E.~Grave, A.~Joulin, and T.~Mikolov.
\newblock Enriching word vectors with subword information.
\newblock \emph{Transactions of the Association for Computational Linguistics},
  5:\penalty0 135--146, 2017.
\newblock ISSN 2307-387X.

\bibitem[Conneau et~al.(2017)Conneau, Lample, Ranzato, Denoyer, and
  J{\'e}gou]{conneau2017word}
A.~Conneau, G.~Lample, M.~Ranzato, L.~Denoyer, and H.~J{\'e}gou.
\newblock Word translation without parallel data.
\newblock \emph{arXiv preprint arXiv:1710.04087}, 2017.

\bibitem[Gaddy et~al.(2016)Gaddy, Zhang, Barzilay, and Jaakkola]{gaddy2016ten}
D.~M. Gaddy, Y.~Zhang, R.~Barzilay, and T.~S. Jaakkola.
\newblock Ten pairs to tag-multilingual pos tagging via coarse mapping between
  embeddings.
\newblock 2016.

\bibitem[Garg et~al.(2019)Garg, Peitz, Nallasamy, and Paulik]{garg2019jointly}
S.~Garg, S.~Peitz, U.~Nallasamy, and M.~Paulik.
\newblock Jointly learning to align and translate with transformer models.
\newblock \emph{arXiv preprint arXiv:1909.02074}, 2019.

\bibitem[Gretton et~al.(2012)Gretton, Borgwardt, Rasch, Sch{\"o}lkopf, and
  Smola]{gretton2012kernel}
A.~Gretton, K.~M. Borgwardt, M.~J. Rasch, B.~Sch{\"o}lkopf, and A.~Smola.
\newblock A kernel two-sample test.
\newblock \emph{Journal of Machine Learning Research}, 13\penalty0
  (Mar):\penalty0 723--773, 2012.

\bibitem[Jansen(2017)]{jansen2017word}
S.~Jansen.
\newblock Word and phrase translation with word2vec.
\newblock \emph{arXiv preprint arXiv:1705.03127}, 2017.

\bibitem[Koehn(2005)]{koehn2005europarl}
P.~Koehn.
\newblock Europarl: A parallel corpus for statistical machine translation.
\newblock In \emph{MT summit}, volume~5, pages 79--86. Citeseer, 2005.

\bibitem[Koehn and Knowles(2017)]{koehn2017six}
P.~Koehn and R.~Knowles.
\newblock Six challenges for neural machine translation.
\newblock \emph{arXiv preprint arXiv:1706.03872}, 2017.

\bibitem[McInnes et~al.(2018)McInnes, Healy, and Melville]{mcinnes2018umap}
L.~McInnes, J.~Healy, and J.~Melville.
\newblock Umap: Uniform manifold approximation and projection for dimension
  reduction.
\newblock \emph{arXiv preprint arXiv:1802.03426}, 2018.

\bibitem[Mikolov et~al.(2013)Mikolov, Le, and Sutskever]{mikolov2013exploiting}
T.~Mikolov, Q.~V. Le, and I.~Sutskever.
\newblock Exploiting similarities among languages for machine translation.
\newblock \emph{arXiv preprint arXiv:1309.4168}, 2013.

\bibitem[Ruder et~al.(2019)Ruder, Vuli{\'c}, and S{\o}gaard]{ruder2019survey}
S.~Ruder, I.~Vuli{\'c}, and A.~S{\o}gaard.
\newblock A survey of cross-lingual word embedding models.
\newblock \emph{Journal of Artificial Intelligence Research}, 65:\penalty0
  569--631, 2019.

\bibitem[Szegedy et~al.(2015)Szegedy, Vanhoucke, Ioffe, Shlens, and
  Wojna]{szegedy2015rethinking}
C.~Szegedy, V.~Vanhoucke, S.~Ioffe, J.~Shlens, and Z.~Wojna.
\newblock Rethinking the inception architecture for computer vision. 2015.
\newblock \emph{arXiv preprint arXiv:1512.00567}, 2015.

\bibitem[Vaswani et~al.(2017)Vaswani, Shazeer, Parmar, Uszkoreit, Jones, Gomez,
  Kaiser, and Polosukhin]{vaswani2017attention}
A.~Vaswani, N.~Shazeer, N.~Parmar, J.~Uszkoreit, L.~Jones, A.~N. Gomez,
  {\L}.~Kaiser, and I.~Polosukhin.
\newblock Attention is all you need.
\newblock In \emph{Advances in neural information processing systems}, pages
  5998--6008, 2017.

\bibitem[Wang and Mahadevan(2009)]{wang2009general}
C.~Wang and S.~Mahadevan.
\newblock A general framework for manifold alignment.
\newblock In \emph{2009 AAAI Fall Symposium Series}, 2009.

\end{thebibliography}

\newpage

\end{document}